\documentclass[10pt, a4paper]{article}

\usepackage{lrec-coling2024} 

\usepackage{latexsym}
\usepackage{array}
\usepackage{booktabs}
\usepackage{siunitx}
\usepackage{amsmath,amssymb,amsthm}
\usepackage{multirow}

\usepackage{amssymb}
\usepackage{pifont}
\usepackage{mathtools}

\usepackage{enumitem}

\usepackage{graphicx}

\usepackage{colortbl}         
\usepackage{amssymb}       

\usepackage{xcolor}
\usepackage{hyperref}
 \definecolor{darkblue}{rgb}{0, 0, 0.5}
  \hypersetup{colorlinks=true, citecolor=darkblue, linkcolor=darkblue, urlcolor=darkblue}

\usepackage{xstring}

\usepackage{color}

\title{Triples-to-isiXhosa (T2X): Addressing the Challenges of Low-Resource Agglutinative Data-to-Text Generation}
\name{Francois Meyer and Jan Buys\\
  Department of Computer Science \\
  University of Cape Town \\
  \texttt{francois.meyer@uct.ac.za, jbuys@cs.uct.ac.za}}

\address{}

\abstract{
Most data-to-text datasets are for English, so the difficulties of modelling data-to-text for low-resource languages are largely unexplored. In this paper we tackle data-to-text for isiXhosa, which is low-resource and agglutinative. We introduce Triples-to-isiXhosa (T2X), a new dataset based on a subset of WebNLG, which presents a new linguistic context that shifts modelling demands to subword-driven techniques. We also develop an evaluation framework for T2X that measures how accurately generated text describes the data. This enables future users of T2X to go beyond surface-level metrics in evaluation. On the modelling side we explore two classes of methods --- dedicated data-to-text models trained from scratch and pretrained language models (PLMs). We propose a new dedicated architecture aimed at agglutinative data-to-text, the Subword Segmental Pointer Generator (SSPG). It jointly learns to segment words and copy entities, and outperforms existing dedicated models for 2 agglutinative languages (isiXhosa and Finnish). We investigate pretrained solutions for T2X, which reveals that standard PLMs come up short. Fine-tuning machine translation models emerges as the best method overall. These findings underscore the distinct challenge presented by T2X: neither well-established data-to-text architectures nor customary pretrained methodologies prove optimal. We conclude with a qualitative analysis of generation errors and an ablation study.
 \\ \newline  
}

\begin{document}

\maketitleabstract
\section{Introduction}

Data-to-text is the task of transforming structured data (e.g. tables or triples) into text describing or summarising the data \citep{gatt-krahmer-2018-survey}.  It is a valuable natural language generation (NLG) task, as 
it enables the separate evaluation of the text content (\emph{what to say}) and its style (\emph{how to say it}) \citep{wiseman-etal-2017-challenges}. 
The majority of data-to-text datasets are in English or other high-resource languages, so such nuanced NLG evaluation is not possible for most low-resource languages.

Existing data-to-text models are designed for the 
linguistic typology of English. This is evident in that there are no studies on the role of subwords in data-to-text. Subwords are not essential for English data-to-text because English is morphologically simple --- words are adequate units for modelling the complexity of datasets. Many examples in data-to-text datasets are instances of common templates. In English these are word-level templates (see Figure \ref{webnlg_example}(a)). As a result, dedicated models for data-to-text do not apply subword segmentation, operating instead on word sequences \citep{wiseman-etal-2017-challenges, wiseman-etal-2018-learning, shen-etal-2020-neural}.

\begin{figure}[t]
\fbox{\begin{minipage}{21.2em}
\textbf{Data triple:} \\
(South Africa, leaderName, Cyril Ramaphosa) \\
($[$\,\,\,   subject \,\,\, $]$, $[$\,\,\,\,relation\,\,\,\,$]$,  $[$\,\,\,\,\,\,\,\,\,\,   object \,\,\,\,\,\,\,\,\, $]$) \\
\\
\textbf{(a) English text and template:}
		\\
Cyril Ramaphosa is the leader of South Africa\\
		$[$\,\,\,\,\,\,\,\,\,\,   object \,\,\,\,\,\,\,\,\, $]$      is the $[$relation$]$$[$\,\,\,\,   subject \,\,\,\, $]$    \\
  \\
\textbf{(b) isiXhosa text and template:}\\
		uCyril Ramaphosa yinkokheli yoMzantsi Afrika\\
			u$[$\,\,\,\,\,\,\,\,\,    object \,\,\,\,\,\,\,\,\,    $]$      \,yin$[$relation$]$ yo$[$\,\,\,\,\,   subject \,\,\,\,\,\, $]$    
\end{minipage}}
\caption{Example from T2X, showing the need for subword-based data-to-text modelling.} 
\label{webnlg_example}
\end{figure}

This is not feasible for agglutinative languages like isiXhosa, where even simple templates are subword-based (see Figure \ref{webnlg_example}(b)).
The problem is compounded by the data scarcity of isiXhosa -- heldout test sets have high proportions of new words, so subword modelling is essential. 
This paper studies data-to-text for isiXhosa: we create a data-to-text dataset with isiXhosa verbalisations, develop a data-focused evaluation framework, and investigate neural approaches for the task.

IsiXhosa is one of South Africa's 12 official languages with over 8 million L1 speakers and 11 million L2 speakers \citep{eberhard-etal-2019-ethnologue}. It is part of the Nguni languages, a group of related languages that are highly agglutinative and conjunctively written (morphemes are strung together to form long words). 
We present and release Triples-to-isiXhosa (T2X),\footnote{\url{https://github.com/francois-meyer/t2x}} the first data-to-text dataset for any Southern African language. 
It was constructed by manually translating part of the English WebNLG dataset and consists of triples of (subject, relation, object) mapped to descriptive sentences. 


Alongside the release of T2X, we conduct a comprehensive investigation into neural data-to-text methods for low-resource agglutinative languages. We explore two prevailing directions of research: (1) LSTM-based encoder-decoder architectures designed to be trained from scratch for data-to-text 
\citep{wiseman-etal-2017-challenges, wiseman-etal-2018-learning, shen-etal-2020-neural}, and (2) finetuning text-to-text pretrained language models (PLMs) \citep{kale-rastogi-2020-text, nan-etal-2021-dart, ribeiro-etal-2021-investigating}. 

Data-to-text models trained from scratch are designed for word-based templates, which is inadequate for agglutinative languages like isiXhosa. 
We propose the subword segmental pointer generator (SSPG), a new neural model aimed at data-to-text for agglutinative languages.\footnote{Code and trained models available at \url{https://github.com/francois-meyer/sspg}.} 
It jointly learns subword segmentation, copying, and text generation. 
Our model adapts the subword segmental approach of \citet{meyer-buys-2022-subword} for sequence-to-sequence modelling and combines it with a copy mechanism. SSPG learns subword segmentations that optimise data-to-text performance and copies entities directly where possible. 
We also propose unmixed decoding, a new decoding algorithm for generating text with SSPG.

We train SSPG on T2X and Finnish data-to-text \citep{kanerva-2019-hockey-generation} (another agglutinative language).  
On both languages SSPG outperforms baselines trained from scratch: on T2X it improves chrF++ by 2.21 and BLEU by 1.11. These results show that \emph{de facto} models for data-to-text 
are not well suited to the unique challenges posed by T2X.
Our experiments on pretrained models yield similar conclusions. We finetune mT5 \citep{lewis-etal-2020-bart} and Afri-mT5 \citep{adelani-etal-2022-thousand}, but neither surpasses SSPG. We only see gains from pretraining when we turn to the unconventional strategy of finetuning English $\rightarrow$ isiXhosa translation models on T2X. So as in the case of models trained from scratch, well-established approaches to finetuning PLMs are suboptimal for T2X. 

In addition to reporting automatic metrics, we develop an extractive evaluation framework for T2X that measures how accurately models describe data. 
Given output text, our framework estimates how well it describes triple data. This allows us to go beyond surface metrics like BLEU, evaluating the content of generations. We apply this framework to all our models, revealing tradeoffs between model capabilities. Subword segmental models copy entities more accurately, while standard subword models verbalise relations more effectively. Based on these findings, we qualitatively analyse the types of errors made by different models.




\section{Related Work}

\subsection{Neural Data-to-text}


Traditionally data-to-text was framed as a series of subtasks \citep{reiter-dake-1997-building} handled separately through pipeline architectures \citep{mckeown-1992-text}. 
This has been combined with deep learning \citep{puduppully1, puduppully2, castro-ferreira-etal-2019-neural}. In our work we approach data-to-text as a sequence-to-sequence task for fully end-to-end learning.
Such approaches can be categorised into neural architectures trained from scratch and finetuned PLMs.

\textbf{Neural architectures}\,\,\,
Data-to-text is a highly structured NLG task, so there is room for exploiting this by equipping models with task-informed inductive biases. 
\citet{wiseman-etal-2018-learning} do this by inducing latent templates and generating text conditioned on these templates. \citet{shen-etal-2020-neural} model the segmentation of text into fragments aligned with data records. Both models are LSTM-based encoder-decoder models that use attention \citep{bahdanau-etal-2015-neural} to incorporate a pointer generator into their decoder \citep{vinyals-etal-2015-pointer}. This enables them to directly copy data tokens during text generation \citep{see-etal-2017-get}.

\textbf{PLMs}\,\,\,
Finetuning text-to-text PLMs, such as BART \citep{lewis-etal-2020-bart} and T5 \cite{t5}, has produced state-of-the-art results for data-to-text \citep{kale-rastogi-2020-text, nan-etal-2021-dart, ribeiro-etal-2021-investigating}. There are very few multilingual data-to-text datasets, so there has not been much work on finetuning multilingual PLMs like mBART \citep{liu-etal-2020-multilingual-denoising} and mT5 \citep{xue-etal-2021-mt5}. In the instances where this has been tried, such as for the Russian WebNLG dataset \citep{zhou-lampouras-2020-webnlg} and the Czech Restaurant dataset \citep{dusek-jurcicek-2019-neural}, results have been promising \citep{gehrmann-etal-2021-gem}.



\subsection{Subword Segmentation}

Subword segmenters like BPE \citep{sennrich-etal-2016-neural} and ULM \citep{kudo-2018-subword} 
operate separately from models trained on their subwords - they are applied in preprocessing.
In the low-resource setting this leads to inconsistent performance \citep{zhu-etal-2019-importance} and oversegmentation \citep{wang-etal-2021-multi-view, acs-2021-exploring}. 
Similar issues arise for morphologically complex languages \citep{klein-tsarfaty-2020-getting, zhu-etal-2019-systematic}.
These problems can be partially attributed to the separation of subword segmenters and model training. 
If segmenters produce suboptimal subwords, models cannot overcome this given insufficient training data.



Alternatively, segmentation can be cast as a latent variable marginalised over during training \citep{kong-etal-2015-segmental, wang-etal-2017-sequence, sun-deng-2018-unsupervised, kawakami-etal-2019-learning}. This leaves segmentation to the model - it is a learnable parameter for optimising the training objective. This has been used to learn subwords for MT \citep{kreutzer-sokolov-2018-learning, he-etal-2020-dynamic, meyer-buys-2023-subword} and low-resource language modelling \citep{downey-etal-2021-masked, meyer-buys-2022-subword}.

\section{Triples-to-isiXhosa (T2X)} \label{sect4_dataset}

\begin{table}[t]
    \centering
    \begin{tabular}{lccc}
    \toprule
         & Train & Valid & Test \\  
         \midrule
    WebNLG 1-triples & 3 114 & 392 & 388 \\ 
    T2X triples & 2 413 & 391 & 378 \\
    T2X verbalisations & 3 859 & 600 & 888 \\
    \bottomrule
    \end{tabular}
    \caption{T2X dataset statistics}
    \label{tab:t2xstats}
\end{table}

WebNLG \citep{gardent-etal-2017-creating} consists of RDF triples from DBpedia paired with text verbalising the triples. Each example is one or more triples (up to seven) paired with a crowd-sourced verbalisation of one or more sentences. 
Multiple verbalisations are included for a large portion of the examples. 
The dataset has been expanded and translated into Russian, using machine translation and manual post-editing \citep{castro-ferreira-etal-2020-2020}. Recently translations have been released for Maltese, Irish, Breton, and Welsh.\footnote{https://github.com/WebNLG/2023-Challenge}  
These datasets cover smaller subsets of WebNLG with up to 1,665 examples per language (about half the size of T2X).
Another data-to-text dataset, Table-to-Text in African languages (TATA) \citep{gehrmann2022tata}, covers several languages including Swahili, which is a Niger-Congo B language like isiXhosa. 
TATA contains less than a thousand examples per language, and generation requires high-level reasoning about the data; in contrast our dataset primarily focuses on linguistic verbalisation ability. 
We are unaware of existing data-to-text datasets for Southern African languages. 

Our dataset, Triples-to-isiXhosa (T2X), is based on the 1-triples in WebNLG version 2.1.\footnote{https://huggingface.co/datasets/web\_nlg}
The choice to only include examples with single triples was motivated by the goal of obtaining a corpus covering a wide range of domains within the available annotation budget.
The data covers 15 DBPedia categories. 
Three categories (Astronaut, Athlete and WrittenWork) are not included in the training data, only in the validation and test sets. 
Dataset statistics are given in Table \ref{tab:t2xstats} and example data-text pairs from the dataset are provided in Figure \ref{webnlg_example} and Tables \ref{tab:t2x_verbalisations_example} \& \ref{generation_examples}. We publicly release the \href{https://github.com/francois-meyer/t2x}{full dataset} for use by future researchers. 

\paragraph{Annotation}

First language isiXhosa speakers who studied the language at university level were presented with triples and English WebNLG verbalisations, and asked to provide 
isiXhosa translations which reflect the content of the triples while phrasing the translations naturally. Annotators discussed questions arising during the process amongst each other, ensuring consistency among annotations. The verbalisations are relatively short, so translating them is an easy task given the isiXhosa proficiency of our annotators. 

In the training and validation sets, only one isiXhosa verbalisation per triple is given for most domains, while the test set has multiple verbalisations (up to 3) for most examples. Multiple verbalisations correspond to different ways of describing the same data triple, capturing variations in phrasing for more nuanced evaluation.  Equivalent verbalisations usually contain synonyms or different word orderings, as shown in the Table \ref{tab:t2x_verbalisations_example} example.

\begin{table}[h] \small
    \centering
    \begin{tabular}{ll}
    \toprule
    \textbf{Data} &  (Germany, leaderName, Angela Merkel)\\ 
\midrule
     \textbf{Text \#1} & Inkokeli yaseJamani ngu-Angela Merkel. \\
              & \emph{The leader of Germany is Angela Merkel.} \\
     \textbf{Text \#2} & U-Angela Merkel yinkokeli yaseJamani. \\
                & \emph{Angela Merkel is the leader of Germany.} \\
    \bottomrule
    \end{tabular}
    \caption{An example T2X data triple mapped to two isiXhosa verbalisations (\emph{with English translations}). }
    \label{tab:t2x_verbalisations_example}
\end{table}

\paragraph{Task difficulty} T2X maps single triples to isiXhosa verbalisations. Unlike WebNLG, it does not contain examples with multiple triples. In that sense it is a simpler task, not requiring combining information from multiple triples, but in some ways T2X is more challenging than existing datasets. For example, E2E \citep{novikova-etal-2017-e2e} covers one domain (restaurants). Much of the text follows a limited set of templates (e.g “[RESTAURANT NAME] is a [TYPE] restaurant in [AREA]”). T2X covers 15 domains and 286 relation types. In this sense T2X is quite challenging, since it would be difficult to model the dataset with a template-based approach. It requires some degree of generalisation and fluency, which is why learning-based approaches are more suitable.

T2X poses a different type of modelling challenge to English data-to-text datasets, because of the agglutinative nature of the isiXhosa language. 
In isiXhosa, morphemes are the primary units of meaning, so effective subword modelling is crucial.
As shown in Figure \ref{webnlg_example}, the underlying syntactic schemas for isiXhosa data-to-text generations are inherently subword-based. For English, a word-based model would cover most examples. For isiXhosa, a subword-based model is essential for even minimal text generation.

\section{Subword Segmental Pointer Generator (SSPG)}

In addition to benchmarking existing models, we propose SSPG to address the challenges of T2X. 
SSPG adds to the line of work designing models trained from scratch for data-to-text. While PLMs are widely used, dedicated data-to-text architectures remain valuable for low-resource languages with few available high-quality pretrained options. SSPG adapts the LSTM encoder-decoder --- LSTMs are well suited to such low-resource tasks \citep{meyer-buys-2022-subword} and persist as the preferred neural architecture for data-to-text \citep{wiseman-etal-2017-challenges, wiseman-etal-2018-learning, shen-etal-2020-neural}. SSPG extends subword segmental modelling \citep{meyer-buys-2022-subword}, which was proposed as a modelling technique for agglutinative languages.
SSPG simultaneously learns how to (1) map data triples to text, (2) segment text into subwords, and (3) when to copy directly from the data.

\subsection{BPE-based Data Encoder}

The encoder is a standard neural encoder for data-to-text: a bi-LSTM that processes data  as flattened sequences of BPE tokens. 
BPE is applied to the data side of a data-to-text dataset.  For example, the triple (France, currency, Euro) could be represented as the sequence ``<s \_Fra nce s> <r currency r> <o \_Euro o>''. Special tokens delimit the boundaries between subject, relation, and object.
BPE is sufficient for data-side segmentation, since most data-to-text datasets in other languages (T2X, Finnish Hockey, and translated WebNLG variants) have English data records.


\subsection{Subword Segmental Decoder}

The decoder is \emph{subword segmental}, i.e., it jointly models the generation and subword segmentation of the output text. We follow the dynamic programming algorithm for subword segmental sequence-to-sequence training outlined by \citet{meyer-buys-2023-subword}, but modify their Transformer-based model to be LSTM-based and extend it to copy subwords.


\begin{figure}[t]
	\includegraphics[width=\linewidth]{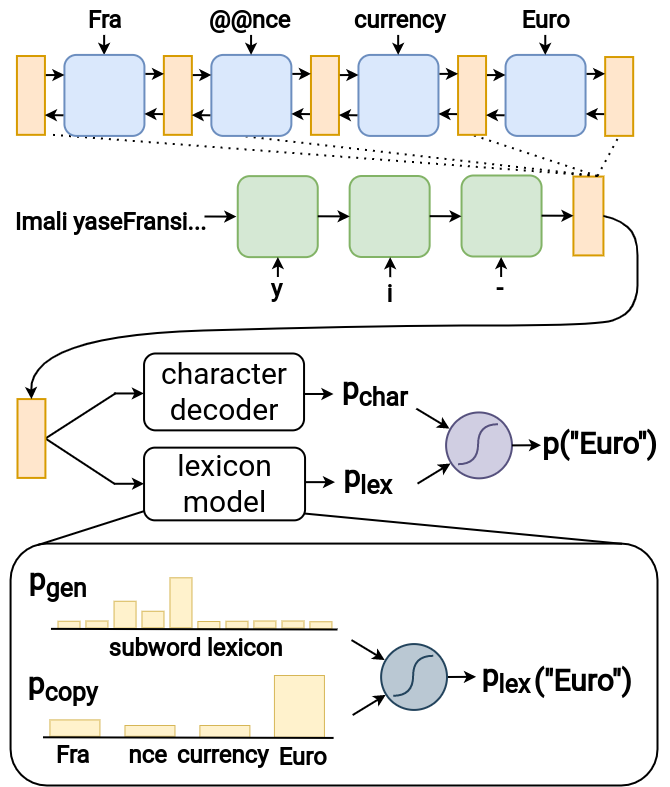}
	\caption{SSPG forward pass for (France, currency, Euro)  $\rightarrow$ ``Imali yaseFransi yi-Euro'' (``The currency of France is the Euro'').
  At each character the next subword probability is computed with a mixture of a character-level decoder and a copy-equipped lexicon model (Eq.~\ref{mixture}).   }
	\label{sspg_architecture}
\end{figure}

During training SSPG processes data-text pairs $(\textbf{x}, \textbf{y})$, where $\textbf{x}$ is a flattened triple of BPE tokens $\mathbf{x} = x_1, x_2, ..., x_{|\mathbf{x}|}$ and $\textbf{y}$ is an unsegmented sequence of characters $\mathbf{y} = y_1, y_2, ..., y_{|\mathbf{y}|}$. 
We compute the probability of the output text conditioned on the input data as	
\begin{align} \label{marginal} 
p(\mathbf{y} | \mathbf{x}) = \sum_{\mathbf{s}: \pi(\mathbf{s}) = \mathbf{y}} p(\mathbf{s} | \mathbf{x}),
\end{align}
where $\mathbf{s}$ is a sequence of subwords and $\pi$ concatenates a sequence of subwords into its pre-segmented character sequence. Therefore we are marginalising over all possible subword segmentations of the output text $\mathbf{y}$.

We model the probability of each subword sequence $\mathbf{s}$ with the chain rule, where each subword probability is computed with a mixture of a character-level decoder and a lexicon model as
\begin{align} \label{mixture}
p(s_{i} | \mathbf{y_{<j}}, \mathbf{x}) =\,\, & g_j p_{\mathrm{char}} (s_{i} | \mathbf{y_{<j}}, \mathbf{x}) + \nonumber\\
& (1-g_j) p_{\mathrm{lex}} (s_{i} | \mathbf{y_{<j}}, \mathbf{x}),
\end{align}
where $\mathbf{y_{<j}}$ is the character sequence up to the character immediately preceding the next subword $s_i$. The character model $p_{\mathrm{char}}$ is a character-level LSTM and the lexicon model $p_{\mathrm{lex}}$ is a softmax distribution over the subword lexicon, which consists of the top $|V|$ (a hyperparameter) most frequent character n-grams in training corpus. The coefficient $g_j$ is computed by a fully connected layer. As shown in Figure \ref{sspg_architecture}, decoder probabilities are conditioned on the input data and the text history by passing attention-based decoder output representations to the subword mixture model components.

As shown in Figure \ref{sspg_architecture}, the output text history is encoded with a character-level LSTM to be tractable. This allows us to compute Eq.~\ref{mixture} at every character position in the output text. We extract probabilities for all subsequent subwords up to a specified maximum segment length and use dynamic programming to efficiently compute Eq.~\ref{marginal}, thereby summing over the probabilities of all possible subword segmentations of $\textbf{y}$.

If we train this model to maximise Eq.~\ref{marginal}, it optimises subword segmentation for its data-to-text task. This is the core idea of subword segmental modelling. It is valuable in settings, such as ours, where subwords are important enough to cast subword segmentation as a trainable parameter.

\subsection{Copying Segments}


The model outlined so far jointly learns data-to-text generation and subword segmentation. This could be useful for low-resource, agglutinative languages, and we include it as a baseline called subword segmental decoder (SSD).
However, it is missing an essential component of most data-to-text models: the ability to copy entities from data, as achieved by pointer generator networks. We want to combine the strengths of subword segmental models and pointer generators. 



We achieve this by including a conditional copy mechanism \citep{gulcehre-etal-2016-pointing} in the lexicon model $p_{\mathrm{lex}}$. We introduce a binary latent variable $z_j$ at each character $j$ indicating whether the subsequent subword $s_i$ is copied from data ($z_j = 1$) or generated from the lexicon ($z_j = 0$). We compute the $p_{\mathrm{lex}}$ in Eq.~\ref{mixture} by marginalising out $z_j$ as
\begin{align} \label{copy_mixture}
&p_{\mathrm{lex}}(s_{i} | \mathbf{y_{<j}}, \mathbf{x})   \\
 &=p (z_{j} = 0| \mathbf{y_{<j}}, \mathbf{x}) p_{\mathrm{gen}} (s_{i} | \mathbf{y_{<j}}, \mathbf{x}) \,+ \nonumber\\
 & \,\,\,\,\,\,\,\,p (z_{j} = 1| \mathbf{y_{<j}}, \mathbf{x}) p_{\mathrm{copy}} (s_{i} | \mathbf{y_{<j}}, \mathbf{x}), \nonumber
\end{align}
where $p_{\mathrm{gen}}$ is a softmax layer over the lexicon and $p_{\mathrm{copy}}$ is the attention distribution over the data tokens. The probabilities $p (z_{j} = 0| \mathbf{y_{<j}}, \mathbf{x})$ and $p (z_{j} = 1| \mathbf{y_{<j}}, \mathbf{x})$ can be viewed as mixture model coefficients (similar to $g_j$ in Eq.~\ref{mixture}). They are computed by a fully connected sigmoid layer, allowing SSPG to learn (based on context) when it can rely on the lexicon's generation model and when it should look to the source to copy BPE tokens directly.

\subsection{Unmixed Decoding}

Standard neural models have one subword vocabulary, so beam search compares next-token probabilities from that vocabulary. 
However, subword segmental modelling uses a mixture model (Eq.~\ref{mixture}), so it is not obvious how to use this for decoding.
\citet{meyer-buys-2023-subword} propose dynamic decoding, which combines information from the two mixture components ($p_{\mathrm{char}}$ and $p_{\mathrm{lex}}$ in Eq .~\ref{mixture}). 
We initially used dynamic decoding to generate text with SSPG, but this resulted in weak performance. Our model's validation performance improved drastically when we developed a new decoding algorithm, which we call \emph{unmixed decoding}. 

We first describe the greedy version of the algorithm. SSPG subword probabilities are a mixture of three distributions: the character decoder, lexicon model, and copy mechanism. Unmixed decoding extracts next-subword probabilities from the three distributions \emph{separately} and selects the next subword with the highest separated (\emph{unmixed}) probability overall. At each decoding step we compute the top next-subword probability from each distribution as
\begin{align}
p^*_\mathrm{\textrm{char}} &=\underset{s}{\max} \, g p_{\mathrm{char}} (s | \cdot), \nonumber \\
p^*_\mathrm{\textrm{gen}} &=\underset{s}{\max} \, (1-g) p (z = 0| \cdot) p_{\mathrm{gen}} (s |  \cdot),  \nonumber\\
p^*_\mathrm{\textrm{copy}} &=\underset{s}{\max} \, (1-g) p (z = 1| \cdot) p_{\mathrm{copy}} (s |  \cdot),  \nonumber
\end{align}
where we omit the conditioning variables of Equations \ref{mixture} and \ref{copy_mixture} for simplification. We store the candidate subwords $s^*_\mathrm{\textrm{char}}$, $s^*_\mathrm{\textrm{gen}}$, $s^*_\mathrm{\textrm{copy}}$ corresponding to these probabilities. We then generate the subword corresponding to the highest probability among $p^*_\mathrm{\textrm{char}}$, $p^*_\mathrm{\textrm{gen}}$, $p^*_\mathrm{\textrm{copy}}$. This process is repeated until the next subword is the end-of-sequence token. It is straightforward to combine unmixed decoding with beam search by extracting the top $k$ subword candidates from each mixture component (resulting in $3k$ initial candidates), ranking these probabilities, and continuing with the top $k$ subwords. 

Each subword generated by unmixed decoding is put forward by one of the mixture components.
During training SSPG learns in which contexts it should use the character decoder, generate from the lexicon, or copy a token from source. Unmixed decoding leverages this information during generation. For example, when the model is confident that the next subword should be copied from data, $p^*_\mathrm{\textrm{copy}}$ will be greater than $p^*_\mathrm{\textrm{char}}$  and $p^*_\mathrm{\textrm{gen}}$, so the next subword is generated by the copy mechanism.

\begin{table*}[t] \small
    \centering
	\begin{tabular}{p{0.09\textwidth}p{0.44\textwidth}|p{0.39\textwidth}} 
		\toprule
            Data & (a) (\textbf{South Africa}, capital, \emph{Cape Town})& (b) (\textbf{Christian Panucci}, club, \emph{Inter Milan}) \\
            \midrule
            Ref & Ikomkhulu lo\textbf{Mzantsi Afrika} li\emph{Kapa}. &U\textbf{Christian Panucci} udlalela i-\emph{Inter Milan}.\\
            SSPG &I-\emph{\textcolor{Red}{Cape Town}} likomkhulu lase\textbf{\textcolor{Red}{South Africa}}. &  U\textbf{\textcolor{Green}{Christian Panucci}} udlalela i-\emph{\textcolor{Green}{Inter Milan}}.\\
            PG&U\emph{\textcolor{Red}{Cape Town}} likomkhulu lase-\textbf{\textcolor{Red}{Afrika}}.  & U\textbf{\textcolor{Red}{Christian Puucci}} udlalela i-\emph{\textcolor{Red}{Indter Milan}}.\\
            BPEMT &Ikomkhulu lo\textbf{\textcolor{Green}{Mzantsi Afrika}} yi\emph{\textcolor{Green}{Kapa}}. & U\textbf{\textcolor{Green}{Christian Panucci}} udlalela i-\emph{\textcolor{Green}{Inter Milan}}.\\
	     \midrule
	       Data & (c) (\textbf{Ethiopia}, leaderName,  \emph{Mulatu Teshome})& (d) (\textbf{Canada}, language, \emph{English}) \\
            \midrule
            Ref \#1 & U\emph{Mulatu Teshome} yinkokheli yase-\textbf{Ethiopia}. & Isi\emph{Ngesi} lulwimi oluthethwa e\textbf{Khanada}.\\ 
            Ref \#2 & Igama lenkokheli e-\textbf{Ethiopia} ngu\emph{Mulatu Teshome}. &Ulwimi lwesi\emph{Ngesi} luthethwa e\textbf{Khanada}.\\
            SSPG &U\emph{\textcolor{Green}{Mulatu Teshome}} yinkokeli yase-\textbf{\textcolor{Green}{Ethiopia}}. &  e\textbf{\textcolor{Red}{Canada}} kuthetwa isi\emph{\textcolor{Green}{Ngesi}}.\\
            PG& Inkokeli yase-\textbf{\textcolor{Green}{Ethiopia}} ngu\emph{\textcolor{Green}{Mulatu Teshome}}. & Ulwimi lwesi\emph{\textcolor{Green}{Ngesi}} luthethwa e\textbf{\textcolor{Red}{Canada}}.\\
            BPEMT &U\emph{\textcolor{Green}{Mulatu Teshome}} yinkokeli yase-\textbf{\textcolor{Green}{Ethiopia}}. & Isi\emph{\textcolor{Green}{Ngesi}} lulwimi oluthethwayo e\textbf{\textcolor{Red}{Canada}}.\\
		\bottomrule
	\end{tabular}
        \caption{Examples from T2X with model outputs. Subject verbalisations are \textbf{bold} and object verbalisations \emph{italicised} to show that some entities should be copied directly while others should be translated. \textcolor{Green}{Green} and \textcolor{Red}{red} show correctly and incorrectly generated entities according to our evaluation framework.}
        \label{generation_examples}
 
	
\end{table*}

\section{Experimental Setup} \label{sec5_setup}


\subsection{Models}

We benchmark T2X with existing models and SSPG. Among our baselines, 3 are trained from scratch (PG, NT, SSD) and 5 are finetuned (mT5-base, mt5-large, Afri-mT5-large, BPE MT, SSMT). We tune hyperparameters for all models as detailed in Appendix \ref{appendix_hyperparams}.

\paragraph{\textbf{Pointer generator (PG)}} is an LSTM-based encoder-decoder model with a copy mechanism. This is commonly employed as a data-to-text baseline \citep{wiseman-etal-2017-challenges, kanerva-etal-2019-template}. 

\paragraph{\textbf{Neural templates (NT)}} \citep{wiseman-etal-2018-learning} learn latent templates for text. The LSTM-based decoder uses a hidden semi-markov model to jointly model templates and text generation. 

\paragraph{\textbf{Subword segmental decoder (SSD)}}
is SSPG without a copy mechanism. We use dynamic decoding \citep{meyer-buys-2023-subword} during generation.

\paragraph{\textbf{mT5}} \citep{xue-etal-2021-mt5} is the multilingual version of T5 \citep{t5}, covering 101 languages, including isiXhosa and Finnish. 

\paragraph{\textbf{Afri-mT5-base}} \citep{adelani-etal-2022-thousand} adapts mT5-base for 17 African language (including isiXhosa) through continued pretraining. 

\paragraph{\textbf{Bilingual pretrained MT (BPE MT \& SSMT)}} 
Low-resource languages like isiXhosa are severely underrepresented in massively multilingual pretraining, so finetuning these PLMs does not guarantee good performance. 
Given the absence of existing NLG datasets, no research has investigated the effectiveness of pretraining and finetuning models for isiXhosa text generation.
To further explore pretraining we turn to the only other publicly available pretrained encoder-decoder models for isiXhosa: machine translation (MT) models.
We use bilingual MT models from \citet{meyer-buys-2023-subword} for English $\rightarrow$ isiXhosa and English $\rightarrow$ Finnish, finetuning them on T2X and Finnish Hockey respectively. 
We finetune 2 MT models for each translation direction: a standard BPE-based model and SSMT (subword segmental machine translation), a model designed for agglutinative MT that jointly learns translation and subword segmentation.

\subsection{Evaluation}

\paragraph{Text overlap}
We compute several automatic metrics: BLEU \citep{papineni-etal-2002-bleu}, chrF \citep{popovic-2015-chrf}, chrF++ \citep{popovic-2017-chrf}, NIST \citep{doddington-2002-automatic}, METEOR \citep{lavie-agarwal-2007-meteor}, ROUGE \citep{lin-2004-rouge}, and CIDEr \citep{conf/cvpr/VedantamZP15}. 
Data-to-text presents an opportunity for more interpretable evaluation, such as quantifying how accurately generations reflect data content. To achieve this we adapt the extractive evaluation framework of \citet{wiseman-etal-2017-challenges} for T2X.

\paragraph{Subject and object extraction}

In T2X (as in the other WebNLG translations) the triples are in English. 
Some entities can be directly copied from data (they are the same in English and isiXhosa), but others should be translated to isiXhosa. For example, in Figure \ref{webnlg_example} the name ``Cyril Ramaphosa'' should be copied, but the country ``South Africa'' should be translated to ``Mzantsi Afrika''. 


\begin{table*} 
	
	\centering
	\begin{tabular}{llccccccc} 
		\toprule
		& Model  & chrF++ & chrF & BLEU & NIST & MET & ROU & CID \\
		\midrule
		   \multirow{4}{*}{Trained from scratch} 
& PG	&	46.24	&	51.09	&	18.90	&	4.32	&	19.84	&	37.48	&	1.32	\\
& NT	&	38.00	&	42.28	&	12.02	&	3.43	&	15.97	&	27.00	&	0.85	\\
& SSD	&	44.77	&	49.91	&	16.16	&	4.14	&	19.85	&	34.97	&	1.16	\\
& SSPG	&	\textbf{48.45}	&	\textbf{53.46}	&	\textbf{20.01}	&	\textbf{4.51}	&	\textbf{21.92}	&	\textbf{38.68}	&	\textbf{1.38}	\\
		\midrule		
        
&	mT5-base	&	41.45	&	46.40	&	14.32	&	3.87	&	20.66	&	33.75	&	1.06	\\
&	Afri-mT5-base	&	42.42	&	47.64	&	16.11	&	4.02	&	20.49	&	34.10	&	1.15	\\
Pretrained + finetuned &	mT5-large	&	46.92	&	52.05	&	19.87	&	4.63	&	23.13	&	39.26	&	1.41	\\
&	BPE MT	&	\textbf{\underline{56.05}}	&	\textbf{\underline{61.53}}	&	\textbf{\underline{27.56}}	&	\textbf{\underline{5.62}}	&	\textbf{\underline{27.24}}	&	\textbf{\underline{47.49}}	&	\textbf{\underline{1.88}}	\\
&	SSMT	&	54.01	&	59.44	&	24.19	&	5.33	&	26.13	&	44.34	&	1.61	\\
		\bottomrule
	\end{tabular}
	\caption{T2X test results. Best scores per category are \textbf{bold} and best scores overall are \underline{underlined}.} 	\label{xhosa_automatic}
\end{table*}

\begin{table*}[t] 
	
	\centering
	\begin{tabular}{llccccccc} 
		\toprule
		
		  && \multicolumn{3}{c}{Subject} & \multicolumn{3}{c}{Object} & Rel\\
		\cmidrule(){3-9}
		
		& Model  & P & R & F1 & P & R & F1 & acc\\
		\midrule
\multirow{4}{*}{Trained from scratch}
&	PG	&	71.30	&	63.37	&	67.10	&	\textbf{77.39}	&	74.40	&	75.86	&	\textbf{75.38}	\\
&	NT	&	72.65	&	73.25	&	72.95	&	67.14	&	68.12	&	67.63	&	38.14	\\
&	SSD	&	\textbf{76.01}	&	84.77	&	80.16	&	71.26	&	59.90	&	65.09	&	67.27	\\
&	SSPG	&	74.83	&	\textbf{88.07}	&	\textbf{80.91}	&	75.42	&	\textbf{\underline{85.99}}	&	\textbf{80.36}	&		70.57\\
            														\midrule
&	mT5-base	&	70.27	&	85.60	&	77.18	&	73.79	&	73.43	&	73.61	&	53.45	\\
&	Afri-mT5-base	&	70.90 & 86.18 & 77.80 & 74.07 & 75.47 & 74.77 & 65.01	\\
Pretrained + finetuned &	mT5-large	&	70.78	&	\textbf{\underline{89.71}}	&	79.13	&	75.22	&	82.13	&	78.52	&	69.37	\\
&	BPE MT	&	74.81	&	83.13	&	78.75	&	77.09	&	\textbf{84.54}	&	80.65	&	\textbf{\underline{87.69}}	\\
&	SSMT	&	\textbf{\underline{77.78}}	&	86.42	&	\textbf{\underline{81.87}}	&	\textbf{\underline{83.17}}	&	83.57	&	\textbf{\underline{83.37}}	&	83.78	\\

		\bottomrule
	\end{tabular}
	\caption{T2X extractive results. Best scores per category are \textbf{bold} and best overall are \underline{underlined}.} 	\label{extractive_results}
\end{table*}


%

To correctly verbalise data entities in T2X models have to learn when to copy entities and when to generate isiXhosa translations of entities. Some models \emph{overcopy}, including English words in the output text where translations are required (see SSPG generation in Table \ref{generation_examples}(a)). Other models copy inaccurately, resulting in missing or partially copied entities in the text (see PG generation in Table \ref{generation_examples}(b)). We can quantify this by casting it as an information retrieval problem: does a generation contain the correctly verbalised entities?

	


For each example we check if the subject/object should be directly copied to the output text. We do this by checking if the entity string from the data triple is present in the reference text. If it is, the entity should be directly copied (like ``Ethiopia'' in Table \ref{generation_examples}(c)). If it is not, the entity should be translated (like ``South Africa'' in Table \ref{generation_examples}(a)). 
Each data-to-text example contains a binary decision for the subject and object, i.e., to translate or to copy. We test how well models learn this decision. 

\paragraph{Relation prediction}

We cannot apply extraction to relations, because they cannot be copied from data. We follow \citet{wiseman-etal-2017-challenges} in training a relation prediction model which we use to estimate how well models capture relations. We finetune \emph{AfroXLMR-large} \citep{alabi-etal-2022-adapting} to predict relation types based on reference texts. The model achieves 85\% heldout accuracy, which is high enough for estimating relation verbalisation capabilities. We apply this predictor to test set generations of models. We compare these to the correct relations in the test set data to estimate how accurately models describe relations.

		

\subsection{Finnish Data-to-Text}

To see if our findings generalise to another agglutinative language we perform experiments on Finnish data-to-text (to the best of our knowledge Finnish is the only other agglutinative language with a data-to-text dataset). The Finnish Hockey dataset \citep{kanerva-2019-hockey-generation} is based on articles about ice hockey games. It contains game statistics and text spans that describe the corresponding game event. The data records are more complex than T2X (up to 12 records, depending on event type), but the texts are single sentences. Our models are trained on the 6,159 data records that are aligned with single text spans. 

\begin{table*}[t] 
	
	\centering
	\begin{tabular}{llccccccc} 
		\toprule
		& Model  & chrF++ & chrF & BLEU & NIST & MET & ROUGE & CID \\
		\midrule
		   \multirow{4}{*}{Trained from scratch} 
& PG	&	37.57&37.98&19.24&4.54&22.74&43.41&2.10	\\  
&	NT	&	32.13&33.70&11.95&3.64&19.17&36.18&1.45	\\
&	SSD	&	33.68&33.78&17.03&4.17&21.53&41.48&1.88	\\
&SSPG	&\textbf{40.12}&\textbf{41.00}&\textbf{20.87}&\textbf{4.63}&\textbf{23.97}&\textbf{44.52}&\textbf{2.13}	\\
		\midrule		
        \multirow{4}{*}{Pretrained + finetuned} 
& mT5-base	&	28.18&30.53&8.39&2.71&14.56&28.49&1.10	\\
&	mT5-large	&	32.70&34.22&13.40&3.46&19.48&39.93&1.75	\\
&	BPE MT	&	\textbf{\underline{42.37}} & \textbf{\underline{41.59}}& \textbf{\underline{22.36}}& \textbf{\underline{5.04}}& \textbf{\underline{24.87}}& \textbf{\underline{46.04}}& \textbf{\underline{2.25}}	\\
&	SSMT	&	36.59&38.52&15.54&4.03&21.62&38.99&1.62	\\	
		\bottomrule
	\end{tabular}
	\caption{Finnish test results. Best scores per category are \textbf{bold} and best scores overall are \underline{underlined}.} 	\label{finnish_automatic}
\end{table*}

\section{Results} \label{sec6_results}

\subsection{Automatic metrics}

The automatic metrics for T2X are reported in Table \ref{xhosa_automatic}. SSPG outperforms the other dedicated data-to-text models trained from scratch across all metrics.
We attribute the low scores of NT to the fact that BPE tokens are not ideal units for learning templates (NT is intended to learn word-level templates, but this led to even worse performance on T2X). 
Considering that PG outperforms SSD, learning subword segmentation is secondary to copying in terms of its importance. 
SSPG combines both aspects to achieve strong performance gains over PG, with increases of 2.21 on chrF++ and and 1.11 on BLUE. This establishes SSPG as a valuable dedicated neural architecture for data-to-text with agglutinative languages.

SSPG outperforms both mT5 variants on chrF++ and BLEU. We believe this is because mT5 is not pretrained on sufficient isiXhosa text and is intended primarily for less structured text-to-text tasks. Interestingly, SSPG even outperforms Afri-mT5. Adapted models like Afri-mT5 (e.g. Afro-XLMR-large \citep{alabi-etal-2022-adapting}) are considered strong pretrained baselines for African languages. The fact that it comes up short shows that standard pretrained solutions are not as reliable for low-resource text generation. 
We only see gains from pretraining whem we turn to the unconventional approach of finetuning MT models.
The best model overall is the finetuned English $\rightarrow$ isiXhosa BPE MT model. This shows the value of pretraining + finetuning for this task, but highlights the lack of high-quality PLMs for isiXhosa. 
BPE MT outperforming SSMT shows that learning subword segmentation for this task is only advantagous when paired with a subword copy mechanism (our ablation results in Section \ref{sec_ablation} confirm this).

Table \ref{finnish_automatic} shows the results for the Finnish Hockey dataset. 
The relative performance of models is similar to T2X. SSPG is the best dedicated model, while the finetuned English $\rightarrow$ Finnish BPE MT model is again best overall. Both models outperform the benchmark established by \citet{kanerva-2019-hockey-generation}, with our best model (BPE MT) achieving a BLEU score gain of 2.69 over their PG model.

Based on our results, finetuning a pretrained MT model can be an effective approach to data-to-text modelling for certain languages. For high-resource languages finetuning text-to-text PLMs might yield better results, but for many low-resource languages MT models will be the best pretrained option available.
For extremely low-resource languages with no available MT models, SSPG is the best choice for training a data-to-text model from scratch for agglutinative languages.

\begin{table}[t] \small
	
	\centering
	\begin{tabular}{lll} 
		\toprule

          \multirow{2}{*}{(b)} &  Ref &  UWilliam M.O. Dawson wazalelwa... \\
           & PG  &  \textcolor{Red}{IWilliam} M. O. Dawson wazalelwa... \\
\midrule
  
        \multirow{2}{*}{(a)} & Ref   &  UNorbert Lammert yinkokeli yaseJamani.\\
        & PG  &   \textcolor{Red}{UNorbu} Lammert yinkokeli yaseJamani.\\
        
     \midrule

    \multirow{2}{*}{(c)} &   Ref  &  I-Dublin yinxalenye yeLeinster.\\
    &  PG   &    \textcolor{Red}{IDubler} yinxalenye yeLeinster.\\

		\bottomrule
	\end{tabular}
	\caption{PG output compared to reference texts. \textcolor{Red}{Red} shows where PG fails on subword copying.} 	\label{pg_mistakes}
\end{table}

\begin{table}[t] \small
    \centering
	\begin{tabular}{lcc} 
		\toprule
	     \textbf{Model and decoding algorithm} & \textbf{chrF++} & \textbf{BLEU}\\
        \midrule
          BPE +copy +beam search (PG) & 48.25 & 18.81 \\
          \midrule
          Subword segmental models & & \\
          +copy +unmixed decoding (SSPG) & \textbf{49.41} & \textbf{20.35}\\
          +copy +dynamic decoding & 43.21 & 14.16 \\
          --copy +unmixed decoding  & 47.11 & 16.87\\
          --copy +dynamic decoding (SSD) & 46.84 & 17.54 \\   
		\bottomrule
	\end{tabular}
	\caption{T2X validation scores for ablations.} 	\label{ablation}
\end{table}

\subsection{Extractive evaluation}

Table \ref{extractive_results} reveals a more mixed account of model performance based on the data content of model generations. Among the models trained from scratch, SSPG achieves the highest F1 scores. Its precision is lower than its recall, indicating some overcopying, but not to such an extent that it undermines its F1 scores. A comparison across all the individual precision and recall scores suggests that SSPG balances copying and translating better than the other models trained from scratch --- it learns in which contexts to copy directly and when to translate instead.

PG outperforms SSPG on relations, which is based on descriptive isiXhosa text (e.g. ``yinkokheli'' in Fig. \ref{webnlg_example} means ``is the leader''). The BPE subwords of PG are sufficient for modelling these isiXhosa phrases. 
However, PG struggles with the subword modelling of entities. isiXhosa has many prefixes that indicate grammatical roles (e.g. ``\emph{u}John'' indicates singular personal proper noun). T2X requires attaching these prefixes to entities correctly. A qualitative analysis of generations reveals that PG struggles with this (see Table \ref{pg_mistakes}). 
SSPG does not seem to make these types of mistakes. By jointly modelling subword segmentation and copying, SSPG learns to combine the two when required.

As in the automatic metrics, SSPG outperforms mT5 but is outperformed by the MT models. SSMT achieves the highest F1 scores for subjects and objects, while BPE MT achieves the highest relation accuracy. This reiterates our findings for the models trained from scratch: BPE subwords are sufficient for descriptive isiXhosa phrases, but modelling subword segmentaion allows SSMT to model subword-based changes to entities.

\subsection{Ablation study} \label{sec_ablation}

Table \ref{ablation} shows the effect of different components on performance. 
A subword segmental encoder-decoder without a copy mechanism (SSD) falls well short of a BPE-based pointer generator (PG).
Adding the copy mechanism improves performance but only if we use unmixed decoding, which is crucial for leveraging the copying ability of SSPG.
While dynamic decoding is useful for high-resource tasks like MT, unmixed decoding is better suited for generating text with subword segmental models trained on smaller datasets.

\section{Conclusion}

We have presented Triples-to-isiXhosa (T2X), a new dataset for isiXhosa data-to-text.
In addition we have proposed SSPG, a new neural model designed for agglutinative data-to-text. 
SSPG outperforms other dedicated data-to-text architectures on two agglutinative languages, isiXhosa and Finnish. 
SSPG is a strong data-to-text model for low-resource agglutinative languages without existing high-quality pretrained models.
In the face of such resource scarcity we also explored finetuning bilingual NMT models, which produced the strongest results overall and should be further investigated as an alternative to PLMs. 
We publicly release T2X and our SSPG implementation to facilitate further research on isiXhosa text generation.

\section{Ethics Statement}

The source of the triples in our T2X dataset is the English WebNLG dataset. A large majority of the content covers Western people, places and events. 
The data content might also contain some biased, outdated or factually incorrect statements. 
This bias has a potentially negative impact on data-to-text models for isiXhosa developed based on the dataset, as some biases might be reflected the the model output and models might perform worse on non-Western people, places and events. Nevertheless, due to the limited availability of structured data in isiXhosa we belief that there is still a clear benefit to releasing this dataset to enable further development of data-to-text approaches for languages such as isiXhosa. Our model architecture, in particular through the copy mechanism, supports generalization to named entities different from those in T2X. Most of the relations covered by the triples (and verbalised in the isiXhosa text) are general enough. However, even with a copy mechanism our models sometime hallucinates and generate incorrect entities in the output, which limits the current reliability of the models.    

\section{Limitations}

Our experiments are limited to two datasets for two languages. 
We cannot make claims about how well SSPG will generalise to other languages and differently structured data-to-text tasks. Our results are very similar across isiXhosa and Finnish and the differences in performance between models are substantial enough that we can confidently claim some generalisability, but only in a narrow linguistic context (simple data-to-text datasets for low-resource agglutinative languages). The limitations regarding the T2X dataset are discused in the ethics statement. 

SSPG takes longer to train than PG because of the additional computation required by its dynamic programming algorithm for summing over latent subword segmentations. The increase in training time depends on the model's maximum segment length. 
Our final SSPG model has a maximum segment length of 5 characters and took aprroximately 4 hours to train on a single NVIDIA A100, as opposed to the 20 minutes training required for our final PG model.

\section*{Acknowledgements}
We thank the annotators who assisted with creating the T2X dataset: Bonke Xakatha, Esethu Mahlumba, Happynes Raselabe, Silu Coki, Anovuyo Tshaka, Yanga Matiwane, and Phakamani Ntentema. We also thank Zukile Jama and Wanga Gambushe for discussions. 

This work is based on research supported in part by the National Research Foundation of South Africa (Grant Number: 129850).  
Computations were performed using facilities provided by the University of Cape Town’s ICTS High Performance Computing team: \url{hpc.uct.ac.za}.
Francois Meyer is supported by the Hasso Plattner Institute for Digital Engineering, through the HPI Research School at the University of Cape Town. 


\bibliographystyle{lrec-coling2024-natbib}

\bibliography{lrec-coling2024-example, anthology, custom}

\label{lr:ref}

\bibliographystylelanguageresource{lrec-coling2024-natbib}
\bibliographylanguageresource{languageresource}

\appendix

\begin{table*}[t] 
	
	\centering
	\begin{tabular}{lcccccl} 
		\toprule
		
	Model  & lr & dropout &  batch sz & BPE size & other \\
    
        \midrule

        PG & 1e-3 & 0.3 & 4 &  500 & \\
        NT & 0.5  & 0.3& 4 &500& discrete states: 10, max segment length: 20\\
        SSD2T  &1e-3 & 0.3&4 & 250& lexicon size: 250, max segment length: 5\\
        SSPG & 1e-3 & 0.5&4 & 1k& lexicon size: 1k, max segment length: 5\\
        \midrule
        mT5-base & 1e-4 & 0.1& 8 & 250k &warmup updates: 0, lr scheduler: fixed\\
        mT5-large & 1e-4 & 0.1 & 8 & 250k&warmup updates: 0, lr scheduler: fixed\\
        BPE MT & 1e-4 & 0.3 & 16 & 10k & warmup updates: 500, lr scheduler: inverse sqrt\\
        SSMT & 1e-4  & 0.3& 16 & 5k & warmup updates: 0, lr scheduler: fixed\\
        \bottomrule

        \end{tabular}
        \caption{Hyperparameter settings for our final T2X models, chosen based on validation chrF++ scores. Some hyperparameters are the same for all our models trained from scratch, including the number of LSTM layers in the encoder \& decoder (1) and the size of the embedding \& hidden layers (128).}
        \label{hyperparams}
\end{table*}

\section{Hyperparameters} \label{appendix_hyperparams}

For each model we ran a grid search over varying hyperparameter settings. We chose the final model to evaluate on the test set based on validation chrF++ score, since it is well suited to evaluate morphologically rich languages \citep{popovic-2017-chrf, google-2022-building}.
Some of the hyperparameters are common to all models (learning rate, dropout, batch size, layers, embedding and hidden size). Other hyperparameters are unique to specific models. For NT we tuned the number of discrete states and the maximum length of segments. For SSD2T and SSPG we tuned the maximum segment length and the lexicon size. For BPE MT and SSMT we tuned the number of warmup updates and the learning rate scheduler.

We tried training our baselines without any subword segmentation, as reported in earlier papers for some of our baselines (NT and PG). However, for all our models we observed improved validation performance when BPE was used for subword segmentation before training. For NT and PG we trained BPE with a shared vocabulary on the data-to-text dataset. For SSD2T and SSPG we trained it on the data half of the data-to-text dataset, since we only use BPE to segment the data in these cases. We tuned BPE vocabulary size separately for each baseline over the range [250, 500, 1k, 2k, 5k] (see Table \ref{hyperparams} for final vocabulary sizes). For the pretrained baselines we used their pretrained subword segmenters.

	


\end{document}